\documentclass[10pt,twocolumn,letterpaper]{article}

\usepackage{iccv}
\usepackage{times}
\usepackage{epsfig}
\usepackage{graphicx}
\usepackage{amsmath}
\usepackage{amssymb}
\usepackage{array}
\usepackage{multirow}
\usepackage{epstopdf}

\usepackage[pagebackref=true,breaklinks=true,letterpaper=true,colorlinks,bookmarks=false]{hyperref}
\newcolumntype{I}{!{\vrule width 1pt}}
\newlength\savedwidth
\newcommand\whline{\noalign{\global\savedwidth\arrayrulewidth
                            \global\arrayrulewidth 1.5pt}%
                   \hline
                   \noalign{\global\arrayrulewidth\savedwidth}}

\iccvfinalcopy 

\def\iccvPaperID{5292} 

\ificcvfinal\fi

\begin{document}

\title{Multi-adversarial Faster-RCNN for Unrestricted Object Detection}

\author{Zhenwei He \qquad Lei Zhang\thanks {Corresponding author} \\
        School of Microelectronics and Communication Engineering, Chongqing University\\
        Shazheng street No.174, Shapingba District, Chongqing 400044, China \\
        {\tt\small \{hzw,leizhang\}@cqu.edu.cn}}

\maketitle

\begin{abstract}
Conventional object detection methods essentially suppose that the training and testing data are collected from a restricted target domain with expensive labeling cost. For alleviating the problem of domain dependency and cumbersome labeling, this paper proposes to detect objects in unrestricted environment by leveraging domain knowledge trained from an auxiliary source domain with sufficient labels. Specifically, we propose a multi-adversarial Faster-RCNN (MAF) framework for unrestricted object detection, which inherently addresses domain disparity minimization for domain adaptation in feature representation. The paper merits are in three-fold: 1) With the idea that object detectors often becomes domain incompatible when image distribution resulted domain disparity appears, we propose a hierarchical domain feature alignment module, in which multiple adversarial domain classifier submodules for layer-wise domain feature confusion are designed; 2) An information invariant scale reduction module (SRM) for hierarchical feature map resizing is proposed for promoting the training efficiency of adversarial domain adaptation; 3) In order to improve the domain adaptability, the aggregated proposal features with detection results are feed into a proposed weighted gradient reversal layer (WGRL) for characterizing hard confused domain samples. We evaluate our MAF on unrestricted tasks including Cityscapes, KITTI, Sim10k, etc. and the experiments show the state-of-the-art performance over the existing detectors.
\end{abstract}

\section{Introduction}

Object detection is a computer vision task which draws many researchers' attentions. Inspired by the development of CNN~\cite{he2016deep, Krizhevsky2012ImageNet, simonyan2014very}, object detection has witnessed a great success in recent years~\cite{Girshick2015Fast, liu2016ssd, Ren2015Faster, yolov3}.

\begin{figure}[t]
\begin{center}
   \includegraphics[width=1.0\linewidth]{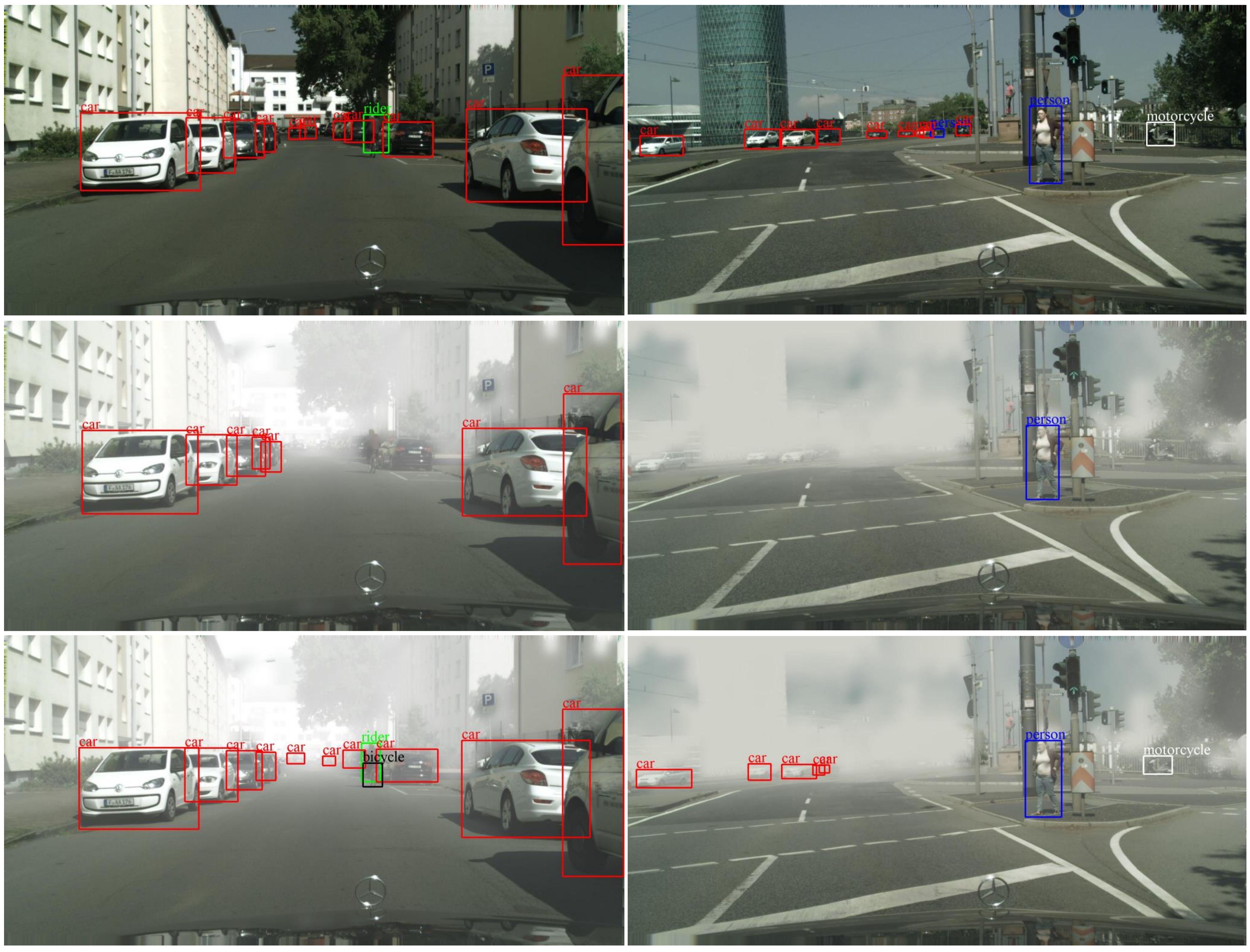}
\end{center}
   \caption{Examples of unrestricted object detection. The first row denotes the pictures from the Cityscapes~\cite{Cordts2016The}, while pictures of the last two rows are detected from the Foggy Cityscapes~\cite{Sakaridis2017Semantic}. The results of the first two rows are detected by the traditional Faster-RCNN~\cite{Ren2015Faster} trained on Cityscapes, and we can see that many objects are missing on the domain shifted Foggy Cityscapes (the second row). The third row shows the results of our approach, and the domain disparity between two datasets can be effectively removed.}
\label{fig:domains}
\end{figure}

Although excellent results have been achieved, object detection in practical application still faces a bottleneck challenge, i.e., detecting objects in the wild where domain shifts always happen. Since the collected datasets~\cite{Chen2015Microsoft, Everingham2015The} are still domain restricted, the trained detectors are difficult to adapt to another domain due to the domain discrepancy between the training data and the testing data it will apply to. Most of conventional detection methods do not take into account the domain discrepancy, which leads to a prominent performance degradation in practice. The influence of domain disparity can be observed in the Figure~\ref{fig:domains}, where we train a VGG16 based Faster-RCNN~\cite{Ren2015Faster} with the Cityscapes~\cite{Cordts2016The} and test the model on Foggy Cityscapes~\cite{Sakaridis2017Semantic}. The results in the second row in Figure~\ref{fig:domains} verify our idea that a considerable performance drop with many objects missing when the domain disparity exists.

Generally, it's difficult to quantitatively remove the domain discrepancy, therefore, for addressing unrestricted object detection challenge, we exploit the mind of domain adaptation and transfer learning~\cite{Long2016Unsupervised, pan2010survey, saenko2010adapting,Zhang2016LSDT} in our detector. In our paradigm, we train the detector on the completely unlabeled target domain, by leveraging a semantic related but distribution different source domain with sufficient labels of bounding boxes. In this way, the domain-invariant features can be learned and there is no any annotation cast for target domain. An example of our proposed detector can be observed in Figure~\ref{fig:domains} (the third row), which shows much better performance than the results of the second row with conventional Faster-RCNN model.

Specifically, we propose a \textbf{m}ulti-\textbf{a}dversarial \textbf{F}aster-RCNN detector (MAF) for adversarial domain adaptation for hierarchical domain features and the proposal features. The hierarchical domain features from the convolutional feature maps progressively present the object position information in the whole image. The proposal features extracted in fully-connected layers can better characterize the semantic information of the generated proposals. In our MAF, we propose multiple adversarial submodules for both domain and proposal features alignment. With similar task, Chen \etal~\cite{Chen2018Domain} proposed a domain adaptive Faster-RCNN (DAF) which demonstrate also that the detector was domain incompatible when image-level distribution difference exists. That is, if the domain feature is aligned, the detector will become domain invariant. Inspired by the wonderful Bayesian perspectives in~\cite{Chen2018Domain}, we focus on the hierarchical domain feature alignment module by designing multiple adversarial domain classifier on each block of the convolution layers for minimizing the domain distribution disparity.

In the proposed MAF, we take into account three important aspects: (1) The multiple domain classifier submodules are learnt to discriminatively predict the domain label, while the backbone network is trained to generate domain-invariant features to confuse the classifier. The multiple two-players adversarial games are implemented by gradient reversal layer (GRL)~\cite{Ganin2014Unsupervised} based optimization in an end-to-end training manner. (2) In the hierarchical domain feature alignment module, the large convolutional feature maps formulate large training sets composed of pixel-wise channel features, which significantly slower the training efficiency. To this end, we propose a \textbf{s}cale \textbf{r}eduction \textbf{m}odule (SRM) without domain feature information loss for reducing the scale of the feature maps by increasing channel number in each convolution block. (3) In the proposal feature alignment module, we propose to aggregate the proposal features with the detection results (\ie, classification scores and regression coordinates) during training of the domain classifier. For further confusing hard samples between domains, we propose a weighted gradient reversal layer (WGRL) to down-weight the gradients of easily confused samples and up-weight the gradients of hard confused samples between domains. The contributions of this paper can be summarized as follows:

\begin{itemize}
\item A multi-adversarial Faster-RCNN (MAF) is introduced for unrestricted object detection tasks. Two feature alignment modules on both hierarchical domain features and aggregated proposal features are proposed with multi-adversarial domain classifier submodules for domain adaptive object detector.

\item In adversarial domain classifier submodule, the scale reduction module (SRM) is proposed for down-scaling the feature maps without information loss, and the training efficiency of our MAF detector is improved.

\item In the aggregated proposal feature alignment module, for improving the domain confusion of proposals, we propose a weighted gradient reversal layer (WGRL) which penalizes the hard confused samples with larger gradient weights and relax the easily confused samples with smaller gradient weights.

\item Exhaustive experiments on Cityscapes~\cite{Cordts2016The}, KITTI~\cite{Geiger2012CVPR}, SIM10K~\cite{Johnson2016Driving}, etc. for unrestricted object detection tasks, which show the superior performance of our MAF over state-of-the-art detectors.
\end{itemize}

\section{Related Work}

\begin{figure*}[t]
\begin{center}
    \includegraphics[width=1.0\linewidth]{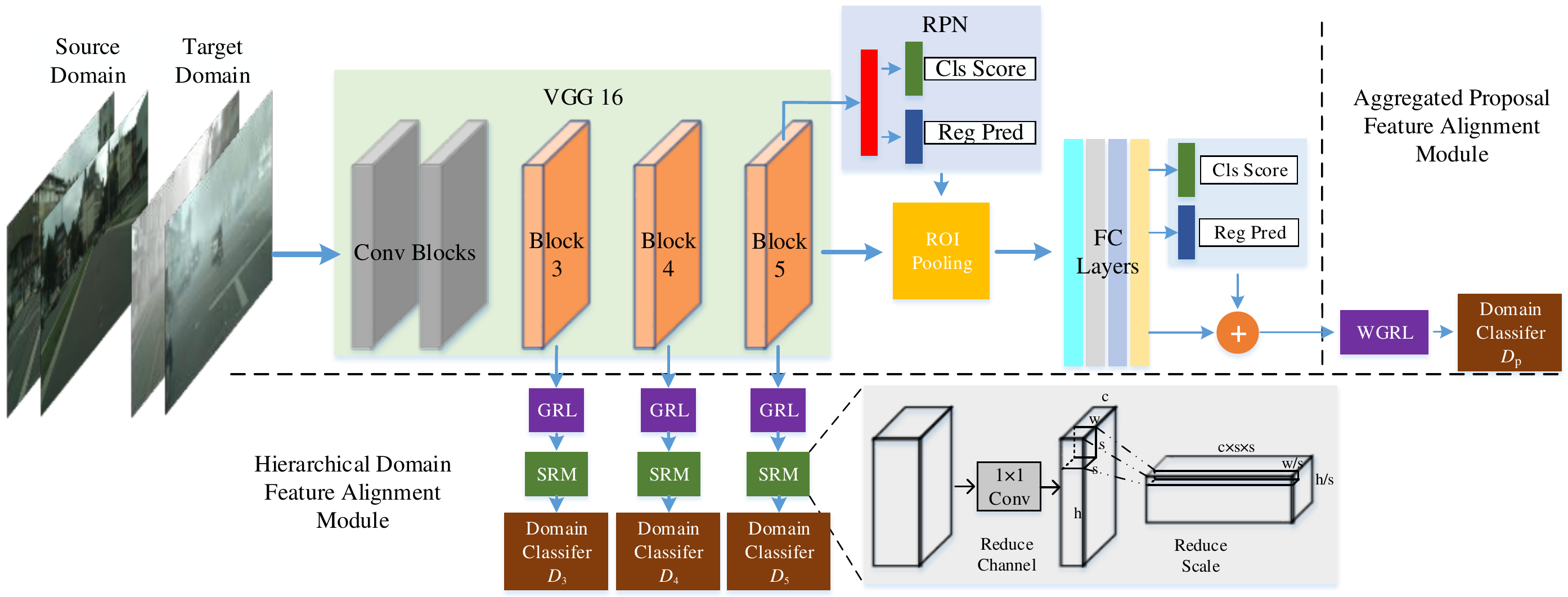}
\end{center}
   \caption{The network structure of our MAF. Inspired by the VGG16 based Faster-RCNN~\cite{Ren2015Faster}, our MAF applies the feature alignment modules on both domain features and proposal features. For the hierarchical domain feature alignment module, multiple adversarial domain classifier submodules are implemented on the block 3,4,5 of the VGG16. GRL layers~\cite{Ganin2014Unsupervised} are used for the adversarial learning strategy and the size of the feature maps are reduced by SRMs. At the proposal feature alignment module, we concatenate the classification scores and bounding box regression results with corresponding features for the domain classifier while the WGRL is introduced for the adversarial learning strategy. SRM is composed of two parts, the first part is a $1\times 1$ convolution layer which is applied to reduce the channel size. After that, a scale reduce part is used to concat $s\times s$ adjacent features, so that the size of the feature maps is reduced.}
\label{fig:netstructure}
\end{figure*}

\textbf{Object Detection.} The object detection is a basic task in computer vision and has been widely studied for many years. The earlier work~\cite{dalal2005histograms,felzenszwalb2010object,Piotr2014Fast} of the object detection were implemented with sliding windows and boost classifiers. Benefited by the success of CNN models~\cite{he2016deep,Krizhevsky2012ImageNet, simonyan2014very}, a number of CNN based object detection methods~\cite{Cai2017Cascade,Fu2017DSSD,Lin2017Focal,Ouyang2017Chained,Shen2017DSOD,Zhang2017Single} have been emerged. The region of interest (ROI) based two-stage object detection methods attracted a lot of attentions in recent years. R-CNN~\cite{girshick2014rich} is the first two-stage detector which classifies the ROIs to find the objects. Girshick \etal~\cite{Girshick2015Fast} further proposed a Fast-RCNN with ROI pooling layer that shares the convolution features, and both the detection speed and accuracy are promoted. After that, Faster-RCNN~\cite{Ren2015Faster} was introduced by Ren \etal, which integrate the Fast-RCNN and Region Proposal Network (RPN) together in an advanced structure. Faster-RCNN further improve the speed and accuracy of detection. In this paper, by taking the Faster-RCNN as backbone, we take into account the mind of domain transfer adaptation for exploring unrestricted object detection task across different domains.

\textbf{Domain Adaptation.} Domain adaptation aims to bridge different domains or tasks by reducing the distribution discrepancy, which has been a focus in various computer vision tasks~\cite{Hoffman2014LSDA,Long2017Conditional,Long2016Unsupervised,wang2018lstn,Zhang2016LSDT}. The domain adaptation has been recently promoted by the powerful feature representation ability of deep learning. Long \etal~\cite{Long2016Unsupervised} implemented the domain adaptation by minimizing the maximum mean discrepancy (MMD) between the two domain-specific fully-connected branches of the CNN. Besides that, domain confusion for feature alignment through two-player game adversarial learning between feature representation and domain classifier motivated by GAN~\cite{Goodfellow2014Generative} was extensive studied in transfer learning~\cite{Li2017MMD,Long2017Conditional,Pei2018Multi,Tzeng2017Simultaneous,zhang2018collaborative}. Tzeng \etal~\cite{Tzeng2017Adversarial} proposed a two-step training scheme to learn a target encoder. Zhang \etal~\cite{zhang2018collaborative} take advantages of several domain classifiers to learn domain informative and domain uninformative features. These works focus on image classification tasks, however, for the object detection task, not only the object categories but also the bounding box location should be predicted, which makes the domain transfer of detectors more challenging. In our MAF detector, the mind of domain adaptation and transfer learning is taken into account for network design, and the adversarial optimization is implemented based on the Gradient Reversal Layer (GRL)~\cite{Ganin2014Unsupervised}. Li \etal~\cite{Li2018Mixed} proposed to transmit the knowledge of the strong categories to the weak categories. In~\cite{Chen2018Domain}, the domain disparity is tackled in both the image level and instance level. However, both works do not fully characterize the hierarchical domain feature alignment and the proposal feature alignment.

\section{The Proposed MAF Detector}
\label{sec3}
In this section, we will introduce our MAF detector. The source domain is marked by $\mathcal{D}_{s}$, and $\mathcal{D}_{t}$ is used for the target domain. In unrestricted setting, the source domain is fully labeled, and $\mathcal{D}_{s}=\left\{(x_{i}^{s}, b_{i}^{s}, y_{i}^{s}) \right\}_{i}^{n_{s}}$ stands for the $n_{s}$ labeled data in the source domain, where the $b_{i}^{s}\in\mathcal{R}^{k\times 4}$ stands for the bounding box coordinates of the $x_{i}^{s}$, and $y_{i}^{s}\in\mathcal{R}^{k\times 1}$ is the category label for corresponding bounding boxes. $\mathcal{D}_{t}=\left\{(x_{j}^{t}) \right\}_{j}^{n_{t}}$ stands for $n_{t}$ completely unlabeled image samples from target domain.

\subsection{Network Structure}
The proposed MAF detector is based on the Faster-RCNN~\cite{Ren2015Faster} framework, and VGG16~\cite{simonyan2014very} with five blocks of the convolution layers is utilized as the backbone of our MAF. The hierarchical domain feature alignment module is implemented on the convolutional feature maps, where the multi-adversarial domain classifier submodules are deployed on blocks 3, 4, and 5. On the top of the network, the aggregated proposal feature alignment module is deployed. With the combination of all feature alignment submodules on both convolution layers and fully collected layer, domain-confused features with domain discrepancy reduced are obtained. It's worth noting that the loss functions including classification loss and smooth L1 loss of the Faster-RCNN are only applied for the source domain. An overview of our network structure is illustrated in Figure~\ref{fig:netstructure}.
Two main modules including 1) hierarchical domain feature alignment module and 2) aggregated proposal feature alignment module formulate the MAF for domain adaptive detection. The former is formulated by multi-adversarial domain classifier submodules, in which a scale reduction module (SRM) is designed on the top of the GRL~\cite{Ganin2014Unsupervised} for down-scaling the feature maps and improving the training efficiency. The latter is formulated by an adversarial domain classifier, in which the aggregated proposal features with detection results are fed as input. For better characterizing the hard-confused samples between domains, the weighted GRL (WGRL) that adaptively re-weights the gradients of easily-confused and hard-confused samples is deployed, which can better improve the adversarial domain adaptation performance.

\subsection{Hierarchical Domain Feature Alignment}
The hierarchical domain feature alignment module aims to calibrate the distribution difference between source and target domain in convolution feature maps, which better characterize the image distribution than semantic layer. A intrinsic assumption is that if the image distribution between domains is similar, the distribution of object-level in the image between domains is basically similar also~\cite{Chen2018Domain}. That is, the distribution difference in the whole image is the primary factor leading to domain discrepancy. In a deep network, the convolutional feature maps in middle level reflect the image information, such as shape, profile, edge, \etc. Therefore, for domain discrepancy minimization between domains, we propose the hierarchical domain feature alignment module which is formulated by multi-adversarial domain classifier submodules in different convolution blocks. The adversarial domain classifier aims to confuse the domain features, with minimax optimization between the domain classifiers and the backbone network. We consider multi-adversarial domain classifiers instead of general single adversarial domain classifier, because hierarchical feature alignment is helpful to the final domain alignment.

Given an image $x_{i}$ from the source domain or target domain, the domain features from the convolution layers of the $m$th block are represented as $C_{m}(x_{i}, w_{m})$, where $w_{m}$ stands for the network parameters. The adversarial classifier submodule at the $m$th block is denoted as $D_{m}$, which is learned to predict the domain label of $x_{i}$. Following the the adversarial learning strategy, the minimax learning of the adversarial classifier submodule in the $m$th convolution block can be written as:
\begin{equation}
\mathop{\min}_{\theta_{m}} \mathop{\max}_{w_{m}} \mathcal{L}_{m}
\label{eq1}
\end{equation}
where $\mathcal{L}_{m}=\sum_{u,v}L_{c}(D_{m}(C_{m}(x_{i}, w_{m})^{(u,v)}, \theta_{m}), d_{i})$, in which $L_{c}$ is the cross entropy loss, the $C_{m}(x_{i}, w_{m})^{(u,v)}$ stands for the channel-wise feature at pixel $(u,v)$ of the feature maps, and $\theta_{m}$ is the domain classifier parameters in the $m$th block. $d_{i}$ is the domain label of sample $x_{i}$, which is labeled as 1 for the source domain and 0 for the target domain. In the Eq.~(\ref{eq1}), the parameters of the backbone network are learned to maximize the cross-entropy loss $L_{c}$, while the parameters of the domain classifier submodule struggle to minimize the loss function. By adversarial learning of the domain classifier with backpropagated gradient reverse (\ie GRL~\cite{Ganin2014Unsupervised}), the feature representations are characterized to be domain-invariant.

In order to efficiently train the hierarchical domain feature alignment module, inspired by~\cite{zhou2018scale}, we introduce a scale reduction module (SRM) which aims at down-scaling the feature maps without information loss. Specifically, SRM contains two steps: 1) A $1\times 1$ convolution layer is implemented to reduce the number of channels of feature maps in each block. This step can achieve domain informative features and reduce the dimensions of domain features for the effective training. 2) Re-align the features by reducing the scale while increasing channel number of the feature maps. This step aims to reduce the size of training set and increase feature dimensionality. In detail, the $s\times s$ adjacent pixels from the feature maps are collected end-to-end to generate a new pixel for the re-shaped feature maps. Obviously, this step is parameterless and easy to compute. The second step is formulated as follows.
\begin{equation}
F_{(u,v,c)}^{S} = F_{(u\times s + c \% s^{2} \% s, v\times s + \lfloor c\% s^{2} / s \rfloor, c / s^{2})}^{L}
\label{eq2}
\end{equation}
where the $F^{L}$ stands for feature maps before the second component. The $(u,v,c)$ presents the element on the $c$th feature map located at $(u,v)$ and count from 0. $F^{S}$ stands for the scale reduced feature maps, and $s$ is the sampling factor, which means the adjacent $s \times s$ pixels of the feature maps are merged into one feature. $\%$ stands for the operation of mod and the $\lfloor . \rfloor$ presents the round down. Since SRM only has parameters in the first component, the number of parameters is reduced while the training efficiency is improved. The two components of our SRM can be clearly observed on the bottom of the Figure~\ref{fig:netstructure}.

\subsection{Aggregated Proposal Feature Alignment}

The object classifier and bounding box regressor trained with the source domain samples can also not be domain adaptive. Therefore, the aggregated proposal feature alignment module aims to achieve semantic alignment while preserving the information for classification and regression. The proposals are obtained from the region proposal network (RPN), which represent the local parts of an image. In order to improve the semantic discriminative of the proposal features, we propose to aggregate the proposal features with the detection results, \ie, classification scores and bounding box regression coordinates, by using concatenation operator. The aggregation brings two kinds of advantages. First, the classification results enrich the information about the categories while the regression results are endowed with position knowledge of the bounding box. Second, the classification and the bounding box regression results improve the discrimination of the features for easily and effectively training the domain classifier.

Given an input image $x_{i}$, the proposal features with respect to the image are represented as $F(x_{i}, w)$, where $w$ is the CNN model parameters. $D_{p}$ is the domain discriminator of the proposal feature alignment module. The loss function of the proposal feature alignment module can be written as
\begin{equation}
 \mathop{\min}_{\theta_{p}} \mathop{\max}_{w} \mathcal{L}_{p}
\label{eq3}
\end{equation}
where $\mathcal{L}_{p}=\sum_{k} {L}_{c}(D_{p}(F^{k}(x_{i}, w)\oplus c^{k} \oplus b^{k}, \theta_{p}), d_{i})$, in which $F^{k}(x_{i}, w)$ is the feature of the $k$th proposal, and $c^{k}$ and $b^{k}$ are the softmax classification scores and the bounding box regression results of the $F^{k}(x_{i}, w)$, respectively. $L_c(\cdot)$ is the cross-entropy loss, $\theta_{p}$ is the domain classifier parameters, and $\oplus$ stands for the concatenation operation.

\begin{figure}[t]
\begin{center}
   \includegraphics[width=1.0\linewidth]{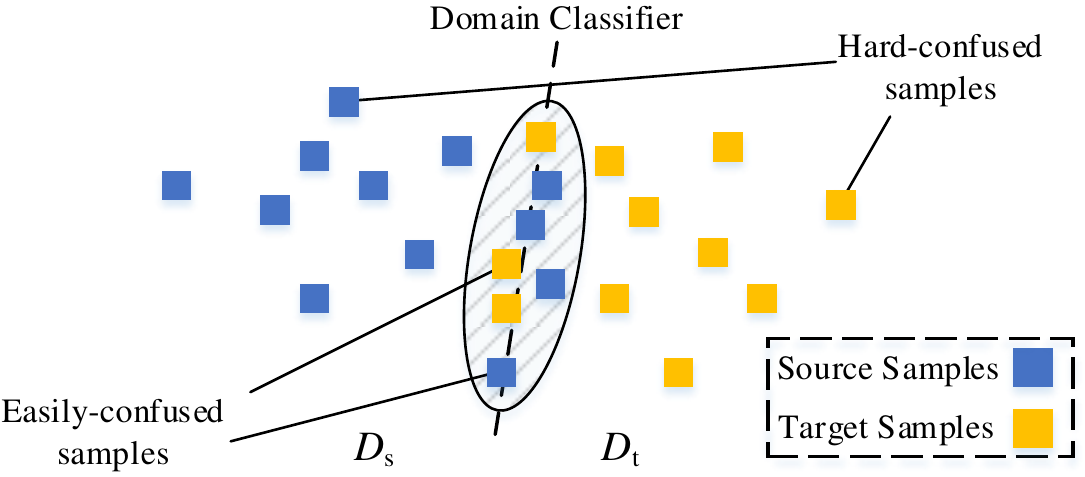}
\end{center}
   \caption{Illustration of WGRL. The blue color stands for the samples from source domain, while the yellow samples stand for the target domain. The samples close to the decision boundary of the domain classifier in the shadow region are recognised as easily-confused samples and up-weighted by our WGRL. The samples outside the shadow region are recognised as hard-confused (\ie, distinguishable) samples that need to be down-weighted.}
\label{fig:wgrl}
\end{figure}

In order to apply the adversarial domain transfer strategy, in the proposal feature alignment module, we propose a weighted gradient reversal layer (WGRL) to relax the easily-confused samples and simultaneously penalize the hard-confused samples, such that better domain confusion can be achieved. An illustration of the proposed WGRL can be viewed in Figure~\ref{fig:wgrl}. The samples close to the domain classifier decision boundary are recognised as the easily-confused samples, \ie, they are not distinguishable by the classifier, while samples far from the decision boundary are hard-confused samples, \ie, the domain discrepancy between these samples in both domains is still large. Thus, we should pay more attention to the distinguishable samples by penalizing these samples with larger weights on their gradients. Specifically, the proposed WGRL regards the scores of the domain classifier as the weights for the corresponding samples. Suppose the probability of one proposal in an image belonging to source domain predicted by the domain classifier to be $p$, the probability belonging to target domain is $1-p$, the gradient before reversal to be $G$, and the gradient after reversal to be $G_{rev}$, then WGRL is written as
\begin{equation}
G_{rev} = -\lambda (d\cdot p + (1-d)(1-p))G
\label{eq4}
\end{equation}
where the $\lambda$ is a hyper-parameter for the WGRL and $d$ is the domain label of the image. According to the Eq. (\ref{eq4}), the predicted scores are used as the weights for the gradients. The higher confidence of the domain classifier means that domain adaptation needs to be further improved, and the samples are automatically up-weighted. Otherwise, those samples with lower confidence of the domain classifier are considered to be indistinguishable and therefore down-weighted. Note that the minus of $-\lambda$ in Eq. (\ref{eq4}) denotes the gradient reverse in optimization.

\subsection{Overview of the MAF Detector}

The overview of our model can be seen in the Figure~\ref{fig:netstructure}. Besides the detection loss $\mathcal{L}_{det}$ of Faster-RCNN, \ie, classification loss and regression loss, our MAF have another two extra minimax loss functions $\mathcal{L}_m$ and $\mathcal{L}_p$, \ie, Eq.(\ref{eq1}) and Eq.(\ref{eq3}) for adversarial domain alignment.

\textbf{Detection loss minimization}. In training of the MAF detector, we utilize the source domain that are full of bounding box labels to train the Faster-RCNN detection loss for the object detection task. The features from the last block of the VGG16~\cite{simonyan2014very} are fed into the RPN to generate a number of proposals for further detection. After that, the ROI pooling layer is used to generate the features with respect to the proposals. The fully-connected layers are trained to get the category labels of the proposals while refining the bounding box coordinates. Note that only the source domain has the annotations for the bounding boxes, the detection loss of the Faster-RCNN is trained on the source domain data.

\textbf{Adversarial domain alignment loss}. The domain alignment loss includes hierarchical domain feature alignment and aggregated proposal feature alignment, which is optimized in an adversarial manner. By jointly considering the Eq.(\ref{eq1}) and Eq.(\ref{eq3}), the proposed adversarial domain alignment loss in MAF can be written as:
\begin{equation}
\mathcal{L}_{t} = \mathcal{L}_{p} +\sum_{m=3}^{5} \mathcal{L}_{m}
\label{eq5}
\end{equation}

\textbf{Overall loss of MAF detector}. With the combination of the detection loss and domain alignment loss, the final loss function of the proposed MAF detector can be written as:
\begin{equation}
\mathcal{L}_{MAF} = \mathcal{L}_{det} + \alpha \mathcal{L}_{t}
\label{eq6}
\end{equation}
where $\mathcal{L}_{det}$ is the loss of Faster-RCNN~\cite{Ren2015Faster} including softmax loss function and smooth $l_1$ loss~\cite{Girshick2015Fast}, and $\alpha$ is a hyper-parameter between the detection loss and domain adaptation loss. The MAF is trained end-to-end with the Eq. (\ref{eq6}). Standard SGD algorithm is implemented to optimize the parameters of the network.

\section{Experiments}

\begin{table*}[t]
\begin{center}
\begin{tabular}{l|cc|p{1cm}<{\centering}|p{1cm}<{\centering}|p{1cm}<{\centering}|p{1cm}<{\centering}|p{1cm}<{\centering}|p{1cm}<{\centering}|p{1cm}<{\centering}|p{1cm}<{\centering}|c}
\whline
  & \textbf{df.} & \textbf{pf.} & person & rider & car & truck & bus & train & mcycle & bicycle & mAP \\
\hline\hline
Faster-RCNN & $\times$ & $\times$& 17.8 & 23.6 & 27.1 & 11.9 & 23.8 & 9.1 & 14.4 & 22.8 & 18.8 \\
\hline
DAF &$\mathbf{\surd}$ &$\mathbf{\surd}$ & 25.0 & 31.0 & 40.5 & 22.1 & 35.3 & 20.2 & 20.0 & 27.1 & 27.6 \\
\hline
MAF* & $\mathbf{\surd}$& $\mathbf{\surd}$& 25.3 & 36.7 & 41.9 & 23.5 & 38.2 & \textbf{36.4} & 18.3 & 28.0 & 30.9 \\
\hline
\multirow{3}*{MAF} &$\times$ & $\mathbf{\surd}$ & 25.6 & 36.8 & 39.9 & 18.8 & 32.0 & 24.1 & 21.3 & 29.2 & 28.5 \\
\cline{2-12}
~ & $\mathbf{\surd}$ &$\times$ &\textbf{29.0} & 38.8 & \textbf{43.9} & 23.2 & 39.6 & \textbf{36.4} & 26.7 & 31.6 & 33.6 \\
\cline{2-12}
~ & $\mathbf{\surd}$ & $\mathbf{\surd}$ & 28.2 & \textbf{39.5} & \textbf{43.9} & \textbf{23.8} & \textbf{39.9} & 33.3 & \textbf{29.2} & \textbf{33.9} & \textbf{34.0} \\
\hline
\end{tabular}
\end{center}
\caption{Results on the validation set of the Foggy Cityscapes. \textbf{df.} denotes domain feature alignment and \textbf{pf.} denotes proposal feature alignment. MAF* means that only one domain feature alignment in the block 5 and the proposal feature alignment are considered.}
\label{w_table}
\end{table*}

In evaluation, we conduct unrestricted object detection experiments on several datasets including Cityscapes~\cite{Cordts2016The}, Foggy Cityscapes~\cite{Sakaridis2017Semantic}, KITTI~\cite{Geiger2012CVPR} and SIM10K~\cite{Johnson2016Driving}. We compare our results with the state-of-the-art domain adaptive Faster-RCNN ~\cite{Chen2018Domain} that we call DAF in experiments and the standard Faster-RCNN. To the best of our knowledge, DAF is the first work on the similar object detection task.

\subsection{Implementation Details}

The experiments in this paper follow the same setting in~\cite{Chen2018Domain}. The source domain of our experiments is sufficiently annotated with bounding boxes and corresponding categories, while the target domain is completely unlabeled. In order to evaluate the performance of the unrestricted object detection, the testing performance of mean average precision (mAP) on the target domain is compared. The trade-off parameter $\alpha$ in Eq. (\ref{eq6}) is set as 0.1 during the training phase. Besides that, for the detection part, we set the hyper-parameters by following~\cite{Ren2015Faster}. We utilize the ImageNet~\cite{Russakovsky2015ImageNet} pre-trained VGG16 model for the initializing our MAF detector. Our model is trained for 50k iterations with the learning rate 0.001 and dropped to 0.0001 for another 20k iterations. Totally, 70k iterations are trained. The minibatch size is set as 2 and the momentum is set as 0.9.

\subsection{Datasets}
Four datasets including Cityscapes~\cite{Cordts2016The}, Foggy Cityscapes~\cite{Sakaridis2017Semantic}, KITTI~\cite{Geiger2012CVPR} and SIM10K~\cite{Johnson2016Driving} are adopted to evaluate the performance of our approach by following~\cite{Chen2018Domain}. The details of these datasets are provided.

\textbf{Cityscapes}: Cityscapes~\cite{Cordts2016The} is designed to capture high variability of outdoor street scenes from different cities. The dataset is captured in common weather conditions and has 5000 images with dense pixel-level labels. These images are collected from 27 cities in different seasons which includes various scenes. Note that the dataset is not originally collected for the object detection task but semantic segmentation, therefore the bounding boxes were generated by the pixel-level annotations as shown in~\cite{Chen2018Domain}.

\textbf{Foggy Cityscapes}: All the images in the Cityscapes~\cite{Cordts2016The} were collected from good weather, the Foggy Cityscapes~\cite{Sakaridis2017Semantic} are derived from the Cityscapes to simulate the foggy scenes and constitutes the images with the fog weather. The inherited pixel labels from Cityscapes are used for generation bounding boxes in experiments. Some examples of the Cityscapes and Foggy Cityscapes are illustrated in Figure~\ref{fig:domains}.

\textbf{KITTI}: The KITTI~\cite{Geiger2012CVPR} is a dataset produced based on an autonomous driving platform. The images of the dataset are captured in a mid-sized city. Totally 14999 images and 80256 bounding boxes are contained in the dataset for the object detection task. In our experiments, 7481 images in the training set are used for both adaptation and evaluation by following~\cite{Chen2018Domain}.

\textbf{SIM10K}: SIM10K~\cite{Johnson2016Driving} is a simulated dataset generated by the engine of the Grand Theft Auto V (GTA V). This dataset contains 10000 images with 58071 bounding boxes of the car. All images of the SIM10k are used as the source domain for training.

\subsection{Experimental Results}

In this section, we evaluate our approach on different datasets to simulate different domain shift scenes. Specially, we evaluate the influence of weather by the first. After that, SIM10k and Cityscapes are implemented to search the domain disparity of synthetic data and real data. Finally, the domain shift caused by different scenes is explored.

\subsubsection{Detection From Cityscapes to Foggy Cityscapes}
\label{sec4.2.1}

We implemented our approach on the Cityspaces~\cite{Cordts2016The} and Foggy Cityspaces~\cite{Sakaridis2017Semantic} to evaluate our MAF under foggy weather condition. We take the Cityscapes as the source domain and the Foggy Cityscapes as the target domain. The VGG16 based Faster-RCNN~\cite{Ren2015Faster} is implemented as the baseline of the experiments. The DAF~\cite{Chen2018Domain}, as a cross-domain detection method, is implemented as the competitor of our MAF. All categories in the Cityscapes are used for the experiments, including the person, rider, car, truck, bus, train, motorcycle and bicycle. The models are tested on the validation set of the Foggy Cityscapes. The results are shown in Table~\ref{w_table}, where the \textbf{df.} stands for the hierarchical domain feature alignment module and the \textbf{pf.} represents the proposal feature alignment module in all experiments.

According to the Table~\ref{w_table}, our MAF achieves the best results among all compared methods. MAF with both domain and proposal feature alignment modules outperforms the DAF by 6.4\%, which shows the significant effectiveness of our approach. Note that MAF with only proposal feature alignment module (\ie, \textbf{pf.}) achieves 28.5\% in mAP, which also outperforms the DAF and the performance of proposal feature alignment module is testified. Besides that, there are some other interesting conclusions can be observed with the results of the MAF* and our approach with only hierarchical domain feature alignment module used. MAF* is a model which contains only one adversarial domain classifier submodule on the block 5, with the submodules on block 3 and 4 removed. Obviously, multi-adversarial domain classifiers on more blocks of convolution layers can significantly improve the domain adaptation performance for better domain-invariant feature representation. With well-aligned domain features, our model achieves much better results and it also verifies our idea that the image distribution calibration in convolutional feature maps is more important than proposal feature alignment in the ultimate domain alignment for the unrestricted object detection task.

\subsubsection{Detection from Synthetic Data to Real Data}

The SIM10k~\cite{Johnson2016Driving} is a dataset composed of the synthetic data. In this experiment, the SIM10k is used as the source domain, while the Cityscapes is used as the target domain. Note that only the category of car is used for the unrestricted object detection task in the experiment. The results are tested on the validation set of the Cityscapes, which are shown in the Table~\ref{sim10k}.

\begin{table}[h]
\begin{center}
\begin{tabular}{l|cc|p{2cm}<{\centering}}
\whline
~ & \textbf{df.} & \textbf{pf.} & AP of Car \\
\hline\hline
Faster-RCNN & $\times$ & $\times$ & 30.1 \\
\hline
DAF & $\mathbf{\surd}$ & $\mathbf{\surd}$ & 39.0 \\
\hline
\multirow{3}*{MAF} & $\times$ & $\mathbf{\surd}$ & 40.1 \\
~ & $\mathbf{\surd}$ & $\times$ & 40.7 \\
~ & $\mathbf{\surd}$ & $\mathbf{\surd}$ & \textbf{41.1} \\
\hline
\end{tabular}
\end{center}
\caption{The results on validation set of target domain Cityscapes, with the SIM10k as the source domain. The average precision (AP) of car is reported. Our MAF with different feature alignment modules (\textbf{df.} and \textbf{pf.}) added is analyzed in the experiment.}
\label{sim10k}
\end{table}

From the results of the Table~\ref{sim10k}, our MAF obtains the best results by comparing to others. Notably, our MAF under different settings can always achieve better performance than the classic Faster-RCNN~\cite{Ren2015Faster}. Our approach also outperforms the DAF~\cite{Chen2018Domain} by 2.1\% in AP. The superiority of the proposed MAF is fully demonstrated for unrestricted object detection. Also, the proposed hierarchical domain feature alignment (\textbf{df.}) can effectively promote the detection performance.

\subsubsection{Detection from One Scene to Another}

Although the weather conditions are similar between Cityscapes and KITTI, there still exists domain disparity caused by different scenes, such as background, view, resolution, camera, \etc In this experiment, we apply the Cityscapes~\cite{Cordts2016The} and KITTI~\cite{Geiger2012CVPR} as the datasets to study the cross-scene object detection. Specifically, the two datasets are implemented as source domain and target domain, alternately. We implement our MAF, DAF~\cite{Chen2018Domain} and Faster-RCNN~\cite{Ren2015Faster} in this experiment. The AP of car is reported for performance comparison. The results of the experiment is shown in Table~\ref{kitti}.

\begin{table}[h]
\begin{center}
\begin{tabular}{l|cc|p{1.5cm}<{\centering}|p{1.5cm}<{\centering}}
\whline
~ & \textbf{df.} & \textbf{pf.} & K $\to$ C & C $\to$ K \\
\hline\hline
Faster-RCNN & $\times$ & $\times$ & 30.2 & 53.5 \\
\hline
DAF & $\surd$ & $\surd$ & 38.5 & 64.1 \\
\hline
\multirow{3}*{MAF} & $\times$ & $\mathbf{\surd}$ & 38.9 & 69.9\\
~ & $\mathbf{\surd}$ & $\times$ & 39.7 & 71.4 \\
~ & $\mathbf{\surd}$ & $\mathbf{\surd}$ &\textbf{41.0} & \textbf{72.1}\\
\hline
\end{tabular}
\end{center}
\caption{The results of the unrestricted object detection task on the Cityspaces and KITTI. The performances of Cityscapes (C)$\to$KITTI (K) and KITTI (K)$\to$Cityspaces (C) are tested. The AP of car is reported for comparison.}
\label{kitti}
\end{table}

In the Table~\ref{kitti}, the K$\to$C means that the KITTI~\cite{Geiger2012CVPR} is used as the source domain while the Cityscapes~\cite{Cordts2016The} is the target domain and vice versa. Obviously, our MAF model gains the best performance under all conditions. The best performance is 8.1\% higher than state-of-the-art DAF method. At this time, the performance of our MAF has been fully verified from hierarchical domain feature alignment to proposal feature alignment.

\subsection{Analysis of Proposal Feature Alignment}

\begin{table*}[t]
\begin{center}
\begin{tabular}{l|p{1cm}<{\centering}|p{1cm}<{\centering}|p{1cm}<{\centering}|p{1cm}<{\centering}|p{1cm}<{\centering}|p{1cm}<{\centering}|p{1cm}<{\centering}|p{1cm}<{\centering}|c}
\whline
  ~ & person & rider & car & truck & bus & train & mcycle & bicycle & mAP \\
\hline\hline
DAF & 25.0 & 31.0 & 40.5 & 22.1 & 35.3 & 20.2 & \textbf{20.0} & 27.1 & 27.6 \\
\hline
MAF* (w/o \textbf{WGRL}) & 25.4 & 36.2 & 41.4 & 22.1 & 36.9 & 31.8 & 19.9 & 28.8 & 30.3 \\
\hline
MAF* (w/o \textbf{Aggregate}) & \textbf{25.5} & 35.6 & \textbf{42.5} & 20.7 & 38.1 & 31.0 & 19.5 & \textbf{29.0} & 30.2 \\
\hline
MAF* & 25.3 & \textbf{36.7} & 41.9 & \textbf{23.5} & \textbf{38.2} & \textbf{36.4} & 18.3 & 28.0 & \textbf{30.9} \\
\hline
\end{tabular}
\end{center}
\caption{Analysis of the proposal feature alignment module. The w/o \textbf{WGRL} denotes that the standard GRL is used in MAF* and w/o \textbf{Aggregate} denotes that the detection results are not concatenated with the proposal feature.}
\label{obj}
\end{table*}

In this section, we analyze the impact of the aggregated proposal feature and WGRL in the proposal feature alignment module. For fair comparison with DAF~\cite{Chen2018Domain} that used one adversarial domain classifier for image-level adaptation, we also use one adversarial domain classifier in domain feature alignment, \ie the MAF* with three settings. In this analysis, the Cityscapes~\cite{Cordts2016The} is used as the source domain and the Foggy Cityspaces~\cite{Sakaridis2017Semantic} is the target domain, by following the same setting as Section~\ref{sec4.2.1}. The analysis results of the experiments are shown in the Table~\ref{obj}.

In Table~\ref{obj}, the WGRL and proposal feature aggregation can be helpful to the final domain adaptation. The concatenation of the proposal features with the classification scores and regression results brings more semantic information for the proposal features, such that the domain classifier can be easily trained for feature confusion. WGRL assigns different weights for easily-confused and hard-confused samples, such that the model pays more attention to the samples that are hard to be confused and gains better training effect. Also, the combination of aggregated proposal feature and WGRL achieves the best mAP, therefore, the performance of the proposed proposal feature alignment module is testified.

\subsection{Influence of IOU Threshold}

\begin{figure}[t]
\begin{center}
   \includegraphics[width=0.95\linewidth]{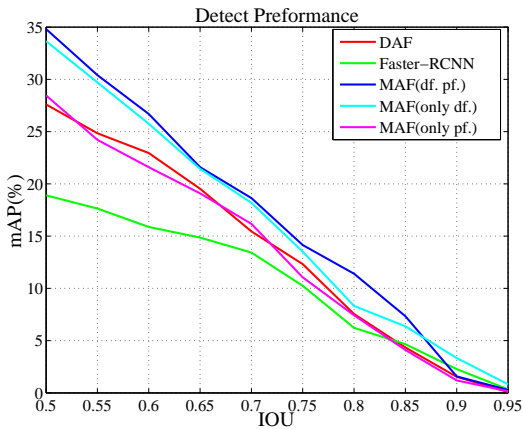}
\end{center}
   \caption{The mAP with different IOU thresholds. MAF, DAF and Faster-RCNN are tested and compared with different IOU thresholds and shown in different colors.}
\label{fig:iou_map}
\end{figure}

The IOU threshold that controls the predicted bounding boxes can also impact the detection results of the testing data. In the previous experiments, the IOU threshold is set as 0.5. In this part, we tune the IOU threshold in the testing phase to study its impact. The Faster-RCNN~\cite{Ren2015Faster}, DAF~\cite{Chen2018Domain}, MAF and MAF with single feature alignment module are implemented with the Cityscapes as source domain and Foggy Cityscapes as target domain. The analysis results of all models are presented in the Figure~\ref{fig:iou_map}.

From Figure~\ref{fig:iou_map}, the mAP drops with the increasing of the IOU threshold for all models. The reason is explicit that a larger IOU threshold means that more predicted bounding boxes are excluded, such that insufficient bounding boxes results in a quick drop of the recall and accuracy. The slope of the curves approximately represents the number of predicted bounding boxes in the corresponding IOU range. Benefit from the multi-adversarial domain adaptation strategy with two feature alignment modules, our MAF achieves the best results under different IOU values. Besides, MAF with only hierarchical feature alignment module, \ie, MAF (only df.) ranks the second place and the importance and effectiveness of the multi-adversarial domain feature alignment is shown. From Figure~\ref{fig:iou_map}, our MAF gets the highest slope on the IOU range 0.8-0.9, the DAF gets the highest slope on range 0.75-0.85, and the Faster-RCNN achieves the highest slope at 0.7-0.8 of IOU range. With the comparison of the slopes, the results reveal that with the domain adaptation, the IOU for unrestricted object detection is increased on the target domain and our MAF with multi-adversarial domain feature alignment achieves the best IOU.

\section{Conclusion}

In this paper, we propose a multi-adversarial Faster-RCNN (MAF) detector for addressing unrestricted object detection problem. Our approach includes two important modules, \ie, hierarchical domain feature alignment and aggregated proposal feature alignment. With an idea that the domain-adaptive object detection depends much on the alignment of image distribution between domains, we therefore propose multi-adversarial domain classifiers in different convolutional blocks for domain confusion of feature maps. For reducing the scale of the feature maps, we propose a SRM for improving the training efficiency of the adversarial domain classifiers. For domain-adaptive detector, we further deploy a proposal feature alignment module by aggregating the detection results for semantic alignment. The aggregated features are feed into the domain classifier with a weighted gradient reversal layer (WGRL), which can automatically focus on the hard confused samples. Our MAF detector can be trained end-to-end by optimizing the domain alignment loss function and the detection loss of Faster-RCNN. We test our model on several datasets with different domains and achieves state-of-the-arts results. The experiments testify the effectiveness of our model.

\noindent\textbf{Acknowledgement:} This work was supported by the National Science Fund of China under
Grants (61771079), Chongqing Youth Talent Program, and the Fundamental
Research Funds of Chongqing (No. cstc2018jcyjAX0250).


{\small
\bibliographystyle{ieee_fullname}
\bibliography{egbib}

\begin{thebibliography}{10}\itemsep=-1pt

\bibitem{Cai2017Cascade}
Zhaowei Cai and Nuno Vasconcelos.
\newblock Cascade r-cnn: Delving into high quality object detection.
\newblock In {\em CVPR}, pages 6154--6162, 2018.

\bibitem{Chen2015Microsoft}
Xinlei Chen, Hao Fang, Tsung-Yi Lin, Ramakrishna Vedantam, Saurabh Gupta, Piotr
  Dollar, and C.Lawrence Zitnick.
\newblock Microsoft coco captions: Data collection and evaluation server.
\newblock {\em Computer Science}, 2015.

\bibitem{Chen2018Domain}
Yuhua Chen, Wen Li, Christos Sakaridis, Dengxin Dai, and Luc~Van Gool.
\newblock Domain adaptive faster r-cnn for object detection in the wild.
\newblock In {\em CVPR}, pages 3339--3348, 2018.

\bibitem{Cordts2016The}
Marius Cordts, Mohamed Omran, Sebastian Ramos, Timo Rehfeld, Markus Enzweiler,
  Rodrigo Benenson, Uwe Franke, Stefan Roth, and Bernt Schiele.
\newblock The cityscapes dataset for semantic urban scene understanding.
\newblock In {\em CVPR}, pages 3213--3223, 2016.

\bibitem{dalal2005histograms}
Navneet Dalal and Bill Triggs.
\newblock Histograms of oriented gradients for human detection.
\newblock In {\em CVPR}, 2005.

\bibitem{Everingham2015The}
Mark Everingham, S.M.Ali Eslami, Luc~Van Gool, Christopher~K.I Williams, John
  Winn, and Andrew Zisserman.
\newblock The pascal visual object classes challenge: A retrospective.
\newblock {\em IJCV}, 111(1):98--136, 2015.

\bibitem{felzenszwalb2010object}
Pedro~F Felzenszwalb, Ross~B Girshick, David McAllester, and Deva Ramanan.
\newblock Object detection with discriminatively trained part-based models.
\newblock {\em IEEE TPAMI}, 32(9):1627--1645, 2010.

\bibitem{Fu2017DSSD}
Cheng-Yang Fu, Wei Liu, Ananth Ranga, Ambrish Tyagi, and Alexander~C Berg.
\newblock Dssd: Deconvolutional single shot detector.
\newblock {\em arXiv preprint arXiv:1701.06659}, 2017.

\bibitem{Ganin2014Unsupervised}
Yaroslav Ganin and Victor Lempitsky.
\newblock Unsupervised domain adaptation by backpropagation.
\newblock {\em arXiv preprint arXiv:1409.7495}, 2014.

\bibitem{Geiger2012CVPR}
Andreas Geiger, Philip Lenz, and Raquel Urtasun.
\newblock Are we ready for autonomous driving? the kitti vision benchmark
  suite.
\newblock In {\em CVPR}, 2012.

\bibitem{Girshick2015Fast}
Ross Girshick.
\newblock Fast r-cnn.
\newblock {\em Computer Science}, 2015.

\bibitem{girshick2014rich}
Ross Girshick, Jeff Donahue, Trevor Darrell, and Jitendra Malik.
\newblock Rich feature hierarchies for accurate object detection and semantic
  segmentation.
\newblock In {\em CVPR}, pages 580--587, 2014.

\bibitem{Goodfellow2014Generative}
Ian. Goodfellow, Jean Pouget-Abadie, Mehdi Mirza, Bing Xu, David Warde-Farley,
  Sherjil Ozair, Aaron Courville, and Yoshua Bengio.
\newblock Generative adversarial nets.
\newblock In {\em NeurIPS}, 2014.

\bibitem{he2016deep}
Kaiming He, Xiangyu Zhang, Shaoqing Ren, and Jian Sun.
\newblock Deep residual learning for image recognition.
\newblock In {\em CVPR}, pages 770--778, 2016.

\bibitem{Hoffman2014LSDA}
Judy Hoffman, Sergio Guadarrama, Eric.S Tzeng, Jeff Donahue, Ross Girshick,
  Trevor Darrell, and Kate Saenko.
\newblock Lsda: Large scale detection through adaptation.
\newblock {\em NeurIPS}, 4:3536--3544, 2014.

\bibitem{Johnson2016Driving}
Matthew Johnson-Roberson, Charles Barto, Rounak Mehta, Sharath~Nittur Sridhar,
  Karl Rosaen, and Ram Vasudevan.
\newblock Driving in the matrix: Can virtual worlds replace human-generated
  annotations for real world tasks?
\newblock {\em arXiv preprint arXiv:1610.01983}, 2016.

\bibitem{Krizhevsky2012ImageNet}
Alex Krizhevsky, Ilya Sutskever, and E.~Geoffrey Hinton.
\newblock Imagenet classification with deep convolutional neural networks.
\newblock In {\em NeurIPS}, pages 1097--1105, 2012.

\bibitem{Li2017MMD}
Chun-Liang Li, Wei-Cheng Chang, Yu Cheng, Yiming Yang, and Barnabas Poczos.
\newblock Mmd gan: Towards deeper understanding of moment matching network.
\newblock In {\em NeurIPS}, pages 2203--2213, 2017.

\bibitem{Li2018Mixed}
Yan Li, Junge Zhang, Kaiqi Huang, and Jianguo Zhang.
\newblock Mixed supervised object detection with robust objectness transfer.
\newblock {\em IEEE TPAMI}, PP(99):1--1, 2018.

\bibitem{Lin2017Focal}
Tsung-Yi Lin, Priya Goyal, Ross Girshick, Kaiming He, and Piotr Dollar.
\newblock Focal loss for dense object detection.
\newblock {\em IEEE TPAMI}, PP(99):2999--3007, 2017.

\bibitem{liu2016ssd}
Wei Liu, Dragomir Anguelov, Dumitru Erhan, Christian Szegedy, Scott Reed,
  Cheng-Yang Fu, and C.~Alexander Berg.
\newblock Ssd: Single shot multibox detector.
\newblock In {\em ECCV}, pages 21--37. Springer, 2016.

\bibitem{Long2017Conditional}
Mingsheng Long, Zhangjie Cao, Jianmin Wang, and Michael~I Jordan.
\newblock Conditional adversarial domain adaptation.
\newblock In {\em NeurIPS}, pages 1640--1650, 2018.

\bibitem{Long2016Unsupervised}
Mingsheng Long, Han Zhu, Jianmin Wang, and Michael~I Jordan.
\newblock Unsupervised domain adaptation with residual transfer networks.
\newblock In {\em NeurIPS}, pages 136--144, 2016.

\bibitem{Ouyang2017Chained}
Wanli Ouyang, Kun Wang, Xin Zhu, and Xiaogang Wang.
\newblock Chained cascade network for object detection.
\newblock In {\em ICCV}, 2017.

\bibitem{pan2010survey}
Sinno~Jialin Pan and Qiang Yang.
\newblock A survey on transfer learning.
\newblock {\em TKDE}, 22(10):1345--1359, 2010.

\bibitem{Pei2018Multi}
Zhongyi Pei, Zhangjie Cao, Mingsheng Long, and Jianmin Wang.
\newblock Multi-adversarial domain adaptation.
\newblock In {\em AAAI}, 2018.

\bibitem{Piotr2014Fast}
Doll¨¢r Piotr, Appel Ron, Belongie Serge, and Perona Pietro.
\newblock Fast feature pyramids for object detection.
\newblock {\em IEEE TPAMI}, 36(8):1532--1545, 2014.

\bibitem{yolov3}
Joseph Redmon and Ali Farhadi.
\newblock Yolov3: An incremental improvement.
\newblock {\em arXiv preprint arXiv:1804.02767}, 2018.

\bibitem{Ren2015Faster}
Shaoqing Ren, Kaiming He, Ross Girshick, and Jian Sun.
\newblock Faster r-cnn: Towards real-time object detection with region proposal
  networks.
\newblock In {\em NeurIPS}, 2015.

\bibitem{Russakovsky2015ImageNet}
Olga Russakovsky, Jia Deng, Hao Su, Jonathan Krause, Sanjeev Satheesh, Sean Ma,
  Zhiheng Huang, Andrej Karpathy, Aditya Khosla, and Michael Bernstein.
\newblock Imagenet large scale visual recognition challenge.
\newblock {\em IJCV}, 115(3):211--252, 2015.

\bibitem{saenko2010adapting}
Kate Saenko, Brian Kulis, Mario Fritz, and Trevor Darrell.
\newblock Adapting visual category models to new domains.
\newblock In {\em ECCV}, pages 213--226. Springer, 2010.

\bibitem{Sakaridis2017Semantic}
Christos Sakaridis, Dengxin Dai, and Luc~Van Gool.
\newblock Semantic foggy scene understanding with synthetic data.
\newblock {\em IJCV}, (11):1--20, 2017.

\bibitem{Shen2017DSOD}
Zhiqiang Shen, Zhuang Liu, Jianguo Li, Yu-Gang Jiang, Yurong Chen, and
  Xiangyang Xue.
\newblock Dsod: Learning deeply supervised object detectors from scratch.
\newblock In {\em CVPR}, pages 1919--1927, 2017.

\bibitem{simonyan2014very}
Karen Simonyan and Andrew Zisserman.
\newblock Very deep convolutional networks for large-scale image recognition.
\newblock {\em arXiv preprint arXiv:1409.1556}, 2014.

\bibitem{Tzeng2017Simultaneous}
Eric Tzeng, Judy Hoffman, Trevor Darrell, and Kate Saenko.
\newblock Simultaneous deep transfer across domains and tasks.
\newblock In {\em ICCV}, 2017.

\bibitem{Tzeng2017Adversarial}
Eric Tzeng, Judy Hoffman, Kate Saenko, and Trevor Darrell.
\newblock Adversarial discriminative domain adaptation.
\newblock In {\em CVPR}, pages 7167--7176, 2017.

\bibitem{wang2018lstn}
Shanshan Wang and Lei Zhang.
\newblock Lstn: Latent subspace transfer network for unsupervised domain
  adaptation.
\newblock In {\em PRCV}, pages 273--284. Springer, 2018.

\bibitem{Zhang2016LSDT}
Lei Zhang, Wangmeng Zuo, and David Zhang.
\newblock Lsdt: Latent sparse domain transfer learning for visual adaptation.
\newblock {\em IEEE TIP}, 25(3):1177--1191, 2016.

\bibitem{Zhang2017Single}
Shifeng Zhang, Longyin Wen, Xiao Bian, Zhen Lei, and Stan~Z Li.
\newblock Single-shot refinement neural network for object detection.
\newblock In {\em CVPR}, pages 4203--4212, 2018.

\bibitem{zhang2018collaborative}
Weichen Zhang, Wanli Ouyang, Wen Li, and Dong Xu.
\newblock Collaborative and adversarial network for unsupervised domain
  adaptation.
\newblock In {\em CVPR}, pages 3801--3809, 2018.

\bibitem{zhou2018scale}
Peng Zhou, Bingbing Ni, Cong Geng, Jianguo Hu, and Yi Xu.
\newblock Scale-transferrable object detection.
\newblock In {\em CVPR}, pages 528--537, 2018.

\end{thebibliography}
}

\end{document}